\documentclass[11pt]{article}
\usepackage[margin=1.05in]{geometry}
\usepackage{amsmath,amssymb,amsthm}
\usepackage{graphicx}
\usepackage{microtype}
\usepackage[colorlinks=true,linkcolor=blue!50!black,citecolor=blue!50!black,urlcolor=blue!50!black]{hyperref}
\usepackage{mathrsfs}
\usepackage{xcolor}
\usepackage[numbers,sort&compress]{natbib}

\newtheorem{definition}{Definition}

\newtheorem{hypothesis}{Hypothesis}
\newcommand{\Imem}{\mathrm{H}}

\title{\bf Abduction Without a Body?\\ \Large Representational Grounding and the Abduction Loop\\ for Scientific Hypothesis Generation}

\author{Michael W. Farmer\\
\small \texttt{farmerm@mac.com}}
\date{August 2026 \quad (v1.18)}

\begin{document}
\maketitle

\begin{abstract}
Can scientific abduction occur without continuous sensorimotor embodiment? Recent arguments in AI and philosophy of science hold that genuine scientific hypothesis generation requires an agent continuously coupled to the physical world through embodied interaction. We defend a narrower and more precise claim: online embodiment is not necessary for every abductive scientific act. Our focus is a specific subclass---\emph{identity abduction}, the inference that two independently developed structures are one object under an explicit correspondence---reached through \emph{representational grounding} rather than bodily interaction. We propose \emph{representational grounding} as an alternative mechanism: an agent may acquire new inferential affordances not through physical interaction but through transformations into representations whose structural organization exposes latent invariants. Scientific diagrams are an especially practical substrate because they embody independently evolved conventions that partially canonicalize symmetry, topology, and operator structure across disciplines---a property we develop as \emph{convention space}, and which yields a concrete answer to a hard retrieval problem: how to find mathematically related work when two fields share no discriminating vocabulary. We operationalize the mechanism as a practical architecture, the \emph{Abduction Loop}: representation generation, motif extraction, convention-space canonicalization, cross-domain retrieval, identity-hypothesis generation, and adversarial deductive verification, with abstention as the designed default. A documented episode---in which a multimodal model, given a figure of a gravitational-memory transport model, generated and then verified the hypothesis that its central differential complex is equivalent to the spherical Kaiser--Squires mass-mapping complex of weak-lensing cosmology---serves as a motivating possibility witness from which the architecture is abstracted, not as evidence of general capability. We close with a falsifiable evaluation program (the DAB-30 benchmark: blinded cross-model runs, visual and textual ablations, adversarial decoys, independent verification). The contribution is a mechanistic proposal, an architecture, and a test program---not an empirical claim that current models possess general scientific creativity.
\end{abstract}

\section{Introduction}

Einstein described scientific discovery as requiring a creative ``Jump'' from experience to explanatory principle---a leap that, he insisted, no logical path compels. Nearly a century later, that Jump has become a central question for artificial intelligence: can machines generate genuinely explanatory hypotheses, or must scientific abduction---Peirce's name for the inference that alone introduces new ideas \citep{peirce1934}---remain grounded in continuous interaction with the physical world? A prominent recent answer says the latter: hypotheses are generated by manipulating mental models grounded in physical sensation \citep{magnani2009}, language models manipulate symbols with no such grounding \citep{harnad1990}, and so they are structurally incapable of the Jump absent embodied world models \citep{zahavy2026,lecun2022}. The recent position crystallizes a longstanding philosophical question rather than originating it: the roles of body, world, and representation in discovery have been contested since Peirce, and the machine case merely makes the question operational. We interpret these arguments as implying a proximate thesis about the reasoning agent, and we address that strongest proximate reading explicitly---if their proponents intend only a developmental or provenance claim (that competence inherits from representations built by embodied communities), then no disagreement remains, and no architectural consequence follows either, since training on the artifacts of embodied communities is what language models already do. Call the proximate premise the Embodiment Necessity Thesis. Its stakes are worth stating plainly: \emph{if the thesis is correct, then no purely representational system---regardless of scale, training, or computational power---should ever produce a genuinely abductive scientific identity.} That is a strong, falsifiable prediction, and this paper develops a mechanistic alternative and an empirical program capable of evaluating it: a documented case that---if accepted as a genuine abductive act---functions as a counterexample, a mechanism explaining how such acts are possible, and a benchmark on which the thesis can lose.

Our question is deliberately narrow: \emph{must every scientific abductive act require continuous online embodiment?} Our answer is no, and our argument is constructive. We describe an alternative grounding mechanism, exhibit a documented episode in which it operated, abstract from that episode a practical system architecture, and specify the experiments that would falsify the proposal.

Four contributions, in order of importance:
(1) a philosophical argument that distinguishes embodiment as a \emph{sufficient} route to grounding from embodiment as a \emph{universally necessary} condition, and refutes only the latter;
(2) a mechanistic account of \emph{representational grounding}---the acquisition of inferential affordances through structure-exposing transformations---with a companion principle of \emph{representational sufficiency};
(3) the concept of \emph{convention space}: independently evolved graphical conventions as a partial canonicalization layer enabling cross-domain retrieval where discriminating lexical overlap is absent; and
(4) the \emph{Abduction Loop}, an architecture that operationalizes the mechanism as divergent candidate generation followed by adversarial verification with abstention as the default, together with the DAB-30 evaluation program that renders the whole proposal falsifiable.

Three scope commitments govern everything that follows. We do not claim that current models possess general scientific creativity; we do not claim that embodiment is unimportant to science or to the historical origin of the very conventions our mechanism exploits; and we do not claim that the motivating episode, taken alone, establishes prevalence, reliability, or autonomy. One documented instance can refute a universal ``cannot''; it can motivate a mechanism and an architecture; it cannot carry more, and we do not ask it to.

\section{The Embodiment Necessity Thesis}

The position we engage deserves its strongest form. Embodied and enactive traditions \citep{varela1991,gibson1979,clark2013} assign the body at least four epistemic roles: \emph{error correction} (the world pushes back on action, disciplining models), \emph{grounding} (symbols acquire content through sensorimotor coupling), \emph{causal intervention} (agents learn structure by manipulation, not observation alone), and \emph{semantic stabilization} (shared embodiment anchors shared reference). In the philosophy of scientific discovery, Magnani's \emph{manipulative abduction} \citep{magnani2009} locates hypothesis generation in the active manipulation of models grounded in physical experience; Einstein's elevator is the canonical exhibit. Applied to machines, the argument runs: language models manipulate symbols without access to their referents \citep{harnad1990}; abduction of new scientific premises requires manipulatively grounded simulation; therefore such models cannot perform the generative Jump, and progress requires physically consistent world models \citep{zahavy2026,lecun2022,bruce2024}.

We take from this argument everything except one premise. Define:

\begin{quote}
\textbf{Embodiment Necessity Thesis (ENT).} Continuous online sensorimotor embodiment of the reasoning agent is necessary for scientific abduction.
\end{quote}

Everything below concerns ENT and only ENT. Embodiment's sufficiency for some abduction, its developmental role in human cognition, and its historical role in creating scientific representations are not in dispute here.

\section{Representational Grounding}

\begin{definition}[Representational grounding]
Representational grounding is the acquisition of new inferential affordances through transformations into representations whose structural organization exposes latent invariants unavailable in the previous representation.
\end{definition}

The definition can be given a computational edge without committing to a formalism. A representation $R_2$ is representationally grounding relative to $R_1$, for an inference class $\mathcal{I}$, if at least one inference in $\mathcal{I}$ is computable from $R_2$ that is not computationally accessible---or is asymptotically more costly---from $R_1$, while the structural invariants required for correctness are preserved. Inferential affordances are thereby measurable in principle: by the set of inferences unlocked, and by the computational cost differential between representations of the same content.

The definition shifts the question. Classical symbol grounding asks what a symbol \emph{refers to} \citep{harnad1990}; representational grounding asks what relations a representation makes \emph{computationally visible}. The two come apart: a coordinate transform refers to nothing new, yet renders a conserved quantity manifest; a phase-space portrait adds no data, yet exposes a limit cycle invisible in a time series; a Penrose diagram compresses an infinite spacetime into a finite figure in which causal structure can be read by eye; Feynman diagrams turn perturbation series into combinatorics; commutative diagrams make compositional identity checkable by path-tracing. In each case the representation is not a passive record of reasoning but a constituent of it---the position long defended in the literatures on diagrammatic reasoning and external cognition \citep{larkin1987,kirsh1994,hutchins1995,zhang1994}, which the machine-abduction debate has largely bypassed by treating ``grounding'' and ``embodiment'' as synonyms. The closest existing account of discovery itself as representational transformation is Nersessian's model-based reasoning program \citep{nersessian2008}: her historical reconstructions---Maxwell's vortex models above all---show conceptual innovation proceeding through the construction and manipulation of intermediary representations, with analogy, imagery, and thought experiment as the transforming operations. Our proposal is continuous with that tradition and indebted to it, with two additions and one departure. We add a computational criterion for when a transformation grounds (the invariant-accessibility condition above) and a retrieval layer her historical cases did not need, since Maxwell carried his source and target domains in one head; convention space is what replaces that single head when the domains live in different communities. The departure concerns interpretation: model-based reasoning is often read, including by Nersessian, as \emph{simulative}---mental modeling with embodied roots---and that reading feeds the ENT. Our claim is that the work her cases attribute to simulation can be relocated to the representations themselves: what the simulation manipulates is what does the grounding, and the manipulation need not be performed by an embodied simulator. The historical evidence, we suggest, underdetermines the two readings; DAB-30 is, among other things, an instrument for separating them.

From the definition follows the principle that carries our argument:

A sufficiency principle follows, which we state plainly rather than as numbered apparatus: \emph{if a representation preserves the structural invariants necessary for an inference while making those invariants computationally accessible, then online embodiment is not a necessary condition for performing that inference.} The principle is conditional, and we are explicit about where its content lies: it identifies the condition under which embodiment ceases to be logically necessary, and the substantive empirical question---whether disembodied systems can in fact \emph{recognize and exploit} such invariants, i.e., whether the antecedent is achievable without a body---is precisely what DAB-30 is designed to evaluate. The principle does not assume that achievability; it locates it as the thing to test. Its force, meanwhile, is that it relocates the operative requirement. What an abductive inference consumes is \emph{access to a representation that exposes the structural invariants on which the inference depends}. Embodiment is one route to such representations; representational transformation is another---and once the requirement is stated at the level of the representation rather than its provenance, the necessity claim has nowhere to attach.

Three notions of grounding should be kept apart in a single breath: \emph{semantic} grounding connects a symbol to its referent; \emph{embodied} grounding connects it to sensorimotor interaction; \emph{representational} grounding connects it to inferentially exploitable structure---and only the third is what an abductive step, as such, consumes. One further clause completes the mechanism: the path from representation to inference runs through \emph{compression}. Scientific representations succeed because they discard irrelevant degrees of freedom while preserving exactly the structural invariants required for explanation; a good diagram is a lossy code optimized for inference.

\begin{definition}[Identity abduction]
Identity abduction is an abductive inference proposing that two apparently distinct mathematical, physical, or computational structures are instances of the same underlying object under an explicit correspondence.
\end{definition}

This is the species our architecture targets, and the definition marks its distance from generic analogy generation: an identity abduction must name its correspondence---which object maps to which, under what restriction---and thereby exposes itself to deductive destruction. An analogy can be suggestive forever; an identity hypothesis is refutable by a single invariant. Because the classification of any episode as identity abduction is exactly what an ENT defender will contest, the definition must be operational, not merely descriptive. An episode qualifies as identity abduction only if four conditions hold: (i) a structural correspondence is proposed that was not explicitly supplied in the prompt or surrounding context; (ii) the correspondence is stated as an explicit mapping between named objects, not a resemblance claim; (iii) the correspondence supports at least one independently verifiable consequence; and (iv) verification is logically downstream of hypothesis generation---the check follows the conjecture, never the reverse. These conditions are scoreable, and DAB-30 scores them; they give the ENT defender something objective to argue with, which is the point. The dialectic thereby moves one level up, deliberately: \emph{the central empirical question is no longer whether an episode occurred, but whether it satisfies an explicit operational definition of identity abduction}---a question a benchmark can answer and a philosophical assertion cannot. Between representation and hypothesis sits the step ENT actually disputes---the recognition step, the registering of a cross-domain invariant match. Verification is not in contention---no one denies machines deduction---so the operational conditions above are designed to isolate the recognition event, condition (i) in particular. We emphasize, finally, that identity abduction is one subclass of scientific abduction, not a proposal about all of it: causal, mechanistic, and interventional hypothesis generation are different inferential animals, and Section~7 states explicitly what the present framework is not expected to capture.

This reframing relocates the debate from how a representation was \emph{acquired}---through a body, through training on the artifacts of embodied communities, through synthesis---to whether the representation \emph{suffices} to support the inference. It also accommodates, rather than opposes, the genetic point that scientific representations are the sediment of embodied practice: convention systems were built by bodies; using them requires none. Proximate mechanism and developmental origin are different questions, and ENT is a claim about the former. ENT, that is, is a claim about who must have the body, not about whether bodies built the library.

\section{Convention Space}

The conceptual centerpiece of this paper is an observation about how science already stores structure. Scientific communities independently evolve graphical conventions that compress domain-specific mathematics into recurring visual motifs: tensor networks and contraction diagrams; causal arrows; bifurcation diagrams; commutative squares; circuit schematics; phylogenetic trees; interaction graphs; field lines; spherical harmonic maps. These conventions are not arbitrary style. They persist because they are \emph{structure-forced}: the drawing that survives is the one whose visual grammar matches the mathematics. A spin-2 field is drawn everywhere as line elements with no distinguished end---a bare segment, or a segment headed identically at both ends---because such a glyph is invariant under $180^\circ$ rotation, and \emph{that invariance is the symmetry}. The two variants are the same motif: what the convention forbids is a \emph{single} arrowhead, because a distinguished end would encode a spin-1 object. The motif class is defined by the invariant, not by the glyph that carries it---which is exactly what treating convention space as a quotient asserts. A differential complex is drawn as boxes and arrows because composition is path-concatenation. A null foliation is drawn as nested cones because causal order is containment.

\begin{definition}[Convention space]
Convention space is the space of graphical motif classes produced by independently evolved scientific drawing conventions that preserve common structural invariants across domains.
\end{definition}

\begin{figure}[t]
\centering
\includegraphics[width=0.97\textwidth]{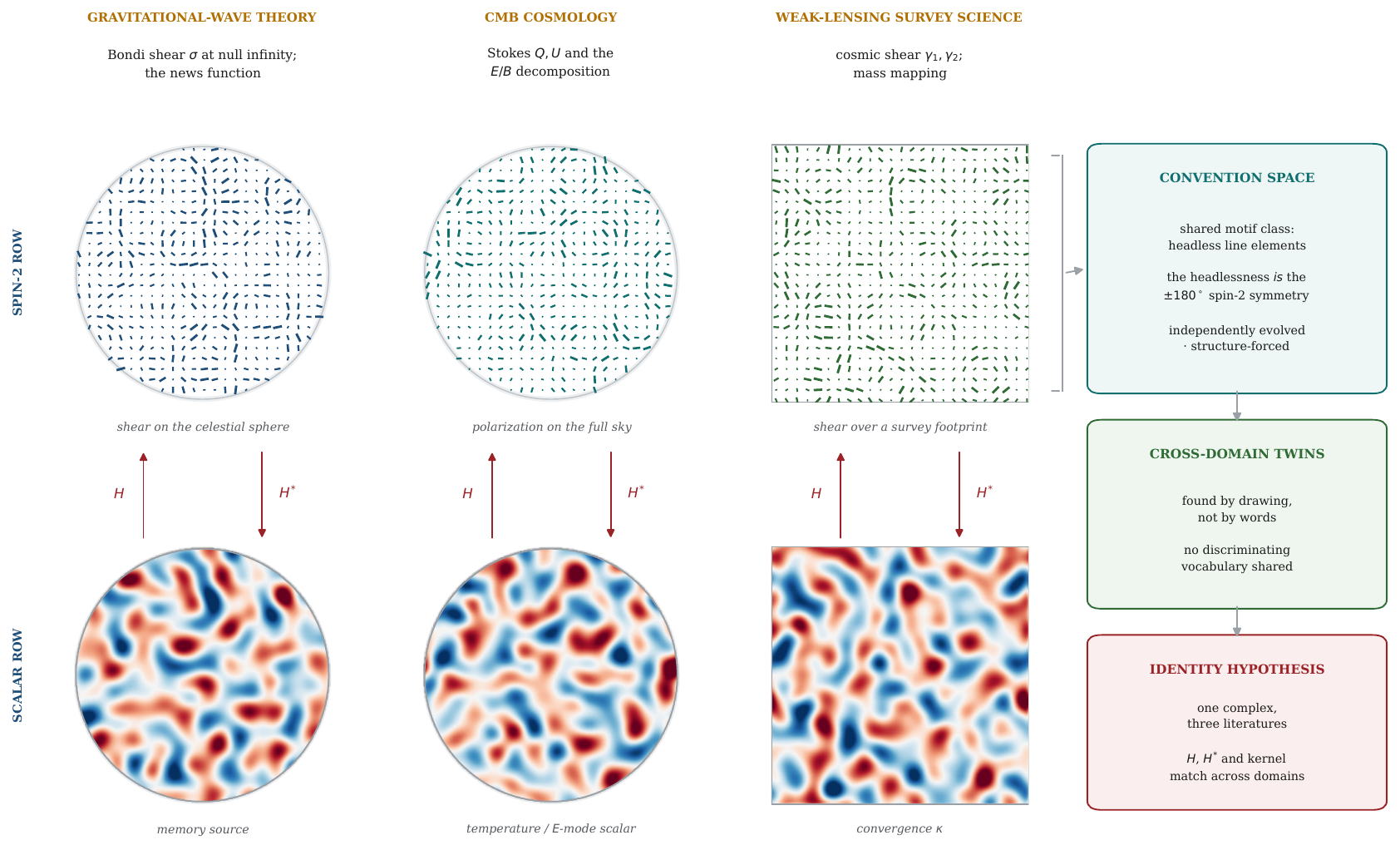}
\caption{Convention space. Three communities that share no discriminating vocabulary---gravitational-wave theory, CMB cosmology, weak-lensing survey science---independently draw the same two-object structure: a spin-2 field of line elements with no distinguished end---drawn bare here, and equivalently as double-headed segments elsewhere, including in the stimulus of Figure~4 (top row) and the scalar potential or source to which it is related by a second-order differential operator (bottom row). Their one common term, ``spin-2,'' localizes nothing; their native terminologies are disjoint; the objects that carry the identification---the complex, its kernel, its spectrum---have no shared name. The shared motif class therefore indexes candidate structural isomorphism, enabling retrieval where discriminating lexical overlap is absent. Panels are synthetic: each column is generated from a single random potential, with the top row its trace-free Hessian and the bottom row its trace, so the drawn relation between rows is the operator relation it depicts. This figure is a schematic of the mechanism, not the input to the motivating case of Section~5.}
\label{fig:conv}
\end{figure}

Diagrams are the first substrate of this theory, not the only one: algebraic normal forms, index-free tensor notation, categorical diagrams, and even standardized sonification exhibit the same convention-forced convergence, and the definition deliberately quantifies over motif classes, not pictures. Disciplinary drawing practices project their mathematics into this space, and the projection is not a courtesy to readers---it is where a community stores what it has decided is essential. Every diagram is a vote about which structure matters; convention space is the tally. Because each community performs this projection independently, and because the projection is forced by structure rather than chosen by taste, convention space partially canonicalizes symmetry, topology, hierarchy, flow, and operator organization \emph{across} communities that share no discriminating vocabulary (Figure~\ref{fig:conv}). That yields a concrete answer to a hard problem in scientific information retrieval---how can a system find mathematically related work when two fields' specific language and notation are disjoint and whatever terms they do share are too coarse to discriminate?---and it is an answer that lexical and even embedding-based text retrieval does not supply, since those inherit the vocabulary gap they are meant to bridge \citep{zhou2018,kang2022}.

A word on what ``no discriminating vocabulary'' means, since the motivating case makes the point concrete. The three literatures of Figure~\ref{fig:conv} are not lexically disjoint in the strict sense: all three would, if pressed, describe their object as a spin-2 field. But that term is non-discriminative---it returns tens of thousands of documents across particle physics, massive gravity, and observational cosmology, and indexes nothing about the structural match that actually mattered. The objects that carried the identification---the two-step differential complex, the $\ell \le 1$ kernel, the $(\ell-1)\ell(\ell+1)(\ell+2)$ spectrum---have no shared name across the three communities, and each community's native terms (Bondi shear and the news function; Stokes parameters and the $E/B$ decomposition; cosmic shear and mass mapping) retrieve only their own literature. The claim throughout is therefore about \emph{discriminating} vocabulary: a shared abstract term is not a bridge, because it does not localize. Convention space localizes.

Why should convergence occur at all? Because different communities are not inventing arbitrary visual languages; they are solving the same optimization problem---encode maximal structural information at minimal cognitive and graphical cost---under the same mathematical constraints. Independent optimization under shared constraints converges: unrelated disciplines arrive at similar diagrams for the same reason unrelated engineers arrive at similar wings and bridges, because the constraints, not the designers, choose the form. Mathematics constrains representation, and convention space is the fossil record of that constraint.

The pattern generalizes far beyond spin-2 fields, and its ubiquity is the argument. Topographic contours exist because level sets of a scalar field are codimension-1 curves that nest and never cross---the drawing gives back the third dimension because the mathematics guarantees it can. Phylogenetic trees exist because descent is a branching topology, so common ancestry is a node and relatedness is path length. Wedge-and-dash stereochemistry exists because chirality is a symmetry no flat drawing can carry, so the convention grew a third dimension to carry it---the wedge \emph{is} the handedness, exactly as the absence of a distinguished end is the spin-2 symmetry. Circuit diagrams exist because conservation laws are node constraints; tensor networks because contraction structure is adjacency. Penrose diagrams exist because conformal compactification makes causal structure a property of position on the page; Feynman diagrams because a perturbation series is combinatorial, so its terms can be drawn and its symmetry factors counted; commutative diagrams because compositional identity is path-equality. Dynkin diagrams are the limiting case: a root system is determined by pairwise angles, so a Lie algebra simply \emph{is} a small labeled graph, with nothing left over. In every case a community converged, without coordination with any other, on a stable graphical grammar \emph{because the mathematics constrained the drawing}---the convention is an equilibrium of a selection process in which representations that misencode structure fail their users and are discarded. Convention space is therefore not a metaphor: it is the union of these evolved grammars, and its canonicalizing power is inherited from the selection pressure that built each one.

\begin{hypothesis}[Convention-space retrieval]
For scientific domains whose graphical conventions preserve mathematical invariants, retrieval in convention space yields higher precision for cross-domain structural correspondences than retrieval based on lexical similarity or generic multimodal embeddings alone.
\end{hypothesis}

The hypothesis is empirical and Section~8 specifies its test; it is the paper's sharpest falsifiable commitment. Formally, convention space can be treated as a quotient space: graphical elements modulo the equivalence ``encodes the same structural invariant,'' with motif classes as its points. This paper remains deliberately agnostic as to its eventual mathematical realization---metric, graph-theoretic, categorical, and learned-latent formulations are all compatible with the mechanism, and choosing among them is an implementation question for the architecture, not a commitment of the theory. The division of labor across the framework is then clean: representational grounding explains \emph{how} a representation enables an inference; convention space explains \emph{where} to search for the representation that enables it; the Abduction Loop explains how to \emph{operationalize} the search; DAB-30 explains how to \emph{test} the whole. Note what it does not say: it does not say visual retrieval is supreme, or that a symbolic-invariant pipeline (fingerprint the operators, search for the fingerprint) could not rival it. It says convention space is a \emph{pre-computed canonical form}---communities have already paid the normalization cost that a symbolic pipeline must reconstruct, and already solved the selection problem of which features matter, since motifs form precisely around the features communities judged essential.

\subsection{Representation versus retrieval}

A deflationary reading of our proposal deserves direct engagement: perhaps the work in the motivating case was done by large-scale multimodal retrieval over an existing literature, and convention space is merely a better index---useful engineering, not a theory of abduction. Two replies. First, the components are jointly necessary and individually insufficient. Retrieval alone returns documents; it does not produce an identity claim, a correspondence dictionary, a verification, or an imported consequence---those are acts of representation and deduction performed on what retrieval returns. Conversely, representation alone cannot discover literature it has never seen; without retrieval there is no candidate to identify. The architecture is the coupling, and attributing its output to either half alone is like attributing an experiment's result to the instrument rather than the hypothesis, or vice versa. Second, ``merely a better index'' concedes the theory's central claim while declining its name: an index is only as good as its canonical form, and the paper's contribution is precisely an account of \emph{why} a canonical form for cross-domain structure already exists---built by selection pressure on independently evolved conventions---and how to search it. That the mechanism, once seen, looks like indexing is not a deflation; canonical forms are what make hard search problems tractable everywhere in computing. The empirical wedge between the readings is Hypothesis~1's ablation: if lexical, shared-abstract-term, or generic-embedding retrieval matches convention-space retrieval on structure-heavy fields, the deflationary reading wins and the theory loses its distinctive content.

\subsection{When convention space should fail}

A mechanism is only as credible as its stated failure conditions, so we say where convention space should \emph{not} work---not as limitations of our evidence, but as predictions of the theory itself. The canonicalizing power of a convention is inherited from selection pressure: representations that misencode structure fail their users and are discarded. Wherever that pressure is absent or misdirected, convention space carries no signal, and the theory predicts failure.

Four such regimes follow directly. \emph{Unselected graphics}: artistic renderings, decorative illustration, and conceptual cartoons are optimized for affect or emphasis, not inference; nothing disciplines their form against the mathematics, so visual similarity among them is evidence of nothing, and retrieval over them should perform at chance. \emph{Stylization drift}: heavily designed figures---press graphics, textbook art directed for beauty---can retain a motif's silhouette while breaking the encoding (a \emph{single} arrowhead added to a spin-2 stick for visual energy destroys the very symmetry the convention exists to carry, while a matched pair at both ends does not); here convention space is actively misleading, worse than chance, because surface similarity survives while structure does not. \emph{Convention collision}: independently evolved grammars sometimes assign one glyph to different invariants---an arrow may denote a vector, a morphism, a causal claim, or mere sequence; a network diagram may encode contraction structure or nothing but adjacency of ideas. Retrieval keyed to the glyph rather than the grammar will propose confident false twins, which is precisely the population our adversarial decoys are constructed from. \emph{Pedagogical topology}: figures drawn to teach often arrange elements by exposition order, salience, or page economics rather than by mathematical structure; their layout reflects the syllabus, not the theory, and the theory predicts they retrieve their curricular neighbors, not their structural ones.

A fifth regime is structural rather than sociological. \emph{Dimensional overflow}: the page has an embedding capacity, and where an invariant's codimension exceeds it, no faithful projection exists. Contours and phylogenetic trees are the benign cases---level sets are intrinsically codimension-1, trees are planar---so the convention discards nothing unrecoverable. Generic phase space is not benign: a four-dimensional flow admits no faithful planar rendering, and communities converge not on one motif but on a family of mutually incompatible projections---Poincar\'e sections trade continuity for dimension, bifurcation diagrams trade the state space for a parameter axis, recurrence plots trade geometry for topology. The theory predicts fragmentation rather than noise: structure survives in each projection, but retrieval keyed to any one of them will miss twins that chose another. The prediction is measurable---cross-domain precision should fall on dynamical-systems figures relative to contour- and tree-class figures, with the loss concentrated in projection mismatch rather than absence of structure.

These failure regimes are not embarrassments to be minimized; they are the theory's second face, and they are testable with the same instrument as its successes. DAB-30's decoy class operationalizes collision and stylization directly, and a well-functioning system's response to all five regimes is the same: abstain. If convention-space retrieval performed \emph{well} on unselected graphics, that would be evidence against our account of why it works at all---the mechanism's failures, correctly located, are part of its confirmation.

\section{A Motivating Case}

We present one documented \emph{design case}---not as empirical validation, but as the source from which the architecture is abstracted and, in its logical role, evidence against the universal necessity claim embodied in ENT. Full protocol details, the epistemic caveats, and the complete transcript commitment appear in Appendix~A. The mathematical content of the equivalence is stated in full below, with conventions and normalization made explicit so that the claim is self-contained. The equivalence is algebraically supported at the level reported here---spectrum, kernel, and normalization---but has not received the independent replication the architecture itself requires, and should be read accordingly rather than as established literature.

On July 10, 2026, during a technical review session, a multimodal LLM (Claude Fable 5, Anthropic) was shown a figure summarizing the CMT-4D model---a null-foliated transport geometry for gravitational-wave memory whose central object is the complex $C^\infty(S^2)/\{\ell\le1\} \xrightarrow{\Imem} \Gamma(\mathrm{STF}_2T^*S^2) \xrightarrow{\Imem^*} C^\infty(S^2)/\{\ell\le1\}$---and given a single instruction: search the web for visually analogous images and examine their sources for possible model enhancements. No target field was named. The model decomposed the figure into motifs, remarked spontaneously that the spin-2 sphere panel follows the drawing convention of CMB polarization and weak-lensing shear maps, retrieved figures from those literatures, and generated the identity hypothesis: the memory complex and the spherical Kaiser--Squires mass-mapping complex \citep{kaiser1993,wallis2022} are equivalent, with shear $\gamma\leftrightarrow\Sigma_{AB}$ and convergence-side scalar $\leftrightarrow$ source.

The hypothesis then survived deduction. Each spin-raising leg $\eth^2$ carries $\sqrt{(\ell+2)!/(\ell-2)!}$ on spin-weighted harmonics; the conventional lensing shear $\gamma=\tfrac12\eth^2\phi$ and the spherical Kaiser--Squires convergence-to-shear kernel add further normalizations, constant and $\ell$-dependent respectively, which is why the correspondence is stated here at the level of the trace-free Hessian complex rather than of any one field's preferred scaling. Composed, that complex carries $\tfrac12(\ell-1)\ell(\ell+1)(\ell+2)$: with the sign convention $\Delta Y_{\ell m}=-\ell(\ell+1)Y_{\ell m}$ throughout, the sphere commutation identity gives $\Imem^*\Imem=\tfrac12\Delta(\Delta+2)$, while the memory operator $\Delta(\Delta+2)$ carries $(\ell-1)\ell(\ell+1)(\ell+2)$ on $\ell\ge2$. The two spectra therefore agree up to the single fixed factor identified below, and the polynomial---not the prefactor---is what the identification turns on. The kernels coincide: the $\ell\le1$ modes unconstrained by shear inversion (whose monopole is related to, though not identical with, the classical mass-sheet degeneracy of convergence reconstruction, and whose dipole is the null mode of $\eth^2$) are precisely the quotient built into the memory complex. The equivalence is claimed as unitary equivalence after electric-parity restriction and realification, modulo normalization (the factor $\tfrac12$ in $\Imem^*\Imem=\tfrac12\Delta(\Delta+2)$ is the realification identity $T_{AB}T^{AB}=2|t|^2$ relating a real symmetric trace-free tensor to its complex spin-2 component, not a free parameter); it is mathematical, not physical---Bondi shear at null infinity and cosmic shear on the sky have disjoint semantics. A disclosure on verification: the initial checks reported here were performed within the same research workflow that generated the hypothesis, and therefore do not satisfy the independence criterion the architecture itself later adopts (Section~6); independent reproduction belongs to the benchmark, not to this case. A minimal operational check confirmed the import: spectral-level Kaiser--Squires inversion recovers a synthetic memory source exactly on $\ell\ge2$ (relative error $1.2\times10^{-16}$), loses injected $\ell\le1$ content as the kernel predicts, and damps noise in proportion to $[(\ell{-}1)\ell(\ell{+}1)(\ell{+}2)]^{-1/2}$, with the prefactor fixed by the tensor and dyad normalization adopted.

The case performs exactly three jobs in this paper: it motivates the mechanism (the operative cue was a drawing convention); it motivates the architecture (every step the model took is a specifiable component); and it motivates the evaluation program (every weakness of the episode---possible contamination, human-supplied strategy, figure-embedded text, single instance---becomes a controlled condition in Section~8). It establishes possibility---and possibility, granted the case's classification as genuinely abductive, is all a counterexample owes.

\section{The Abduction Loop}

\begin{figure}[t]
\centering
\includegraphics[width=0.99\textwidth]{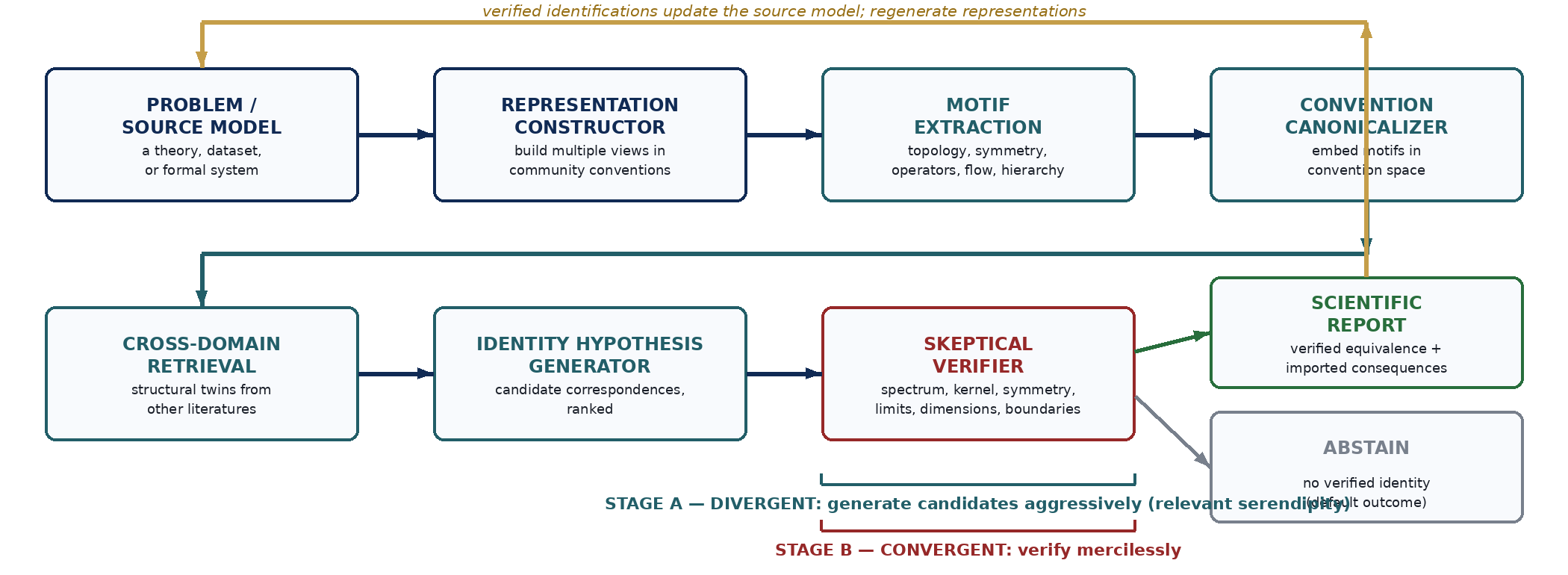}
\caption{The Abduction Loop. Stage A (divergent) generates candidate structural correspondences aggressively; Stage B (convergent) attempts to destroy each one; abstention is the designed default outcome. Verified identifications import mathematics into the source model, whose regenerated representations close the loop.}
\label{fig:arch}
\end{figure}

We now state the architecture abstracted from the case (Figure~\ref{fig:arch}). The mechanism is representational grounding; diagrams are its first substrate; the current diagram-first implementation---the \emph{Diagram Abduction Agent} (DAA)---is Version~1 of the architecture, not the architecture itself.

\subsection{Stages}

\textbf{1. Representation construction.} Retrieval operates over representations rather than raw theories because different representations expose different structural invariants; construction therefore precedes and conditions everything downstream. Build renderings of the source model in multiple representational views drawn in the conventions of established practice---spherical maps, complexes, foliations, networks. Multiplicity matters: each view exposes different invariants, and the generator may itself be a model (in the motivating case, the input figure was AI-generated from the mathematics).

\textbf{2. Motif extraction.} A vision--language component segments each view into motifs and extracts their structural content: topology, symmetry class, hierarchy, flow direction, adjacency, operator organization.

\textbf{3. Convention-space canonicalization and cross-domain retrieval.} Motifs are embedded in convention space and matched against the drawn literature of \emph{other} fields. The target of retrieval is not images but \emph{structural twins}: figures whose motifs plausibly encode the same mathematics. A deep-versus-surface similarity gate filters gestalt matches from structural ones.

\textbf{4. Identity-hypothesis generation.} For each surviving twin, trace to source, extract the underlying object, and produce a candidate correspondence---explicitly a conjectured homomorphism or equivalence, with the proposed dictionary (which object maps to which, under what restriction).

\textbf{5. Adversarial verification.} Attempt to destroy every candidate: spectrum, kernel, symmetry class, operator order, limiting behavior, boundary conditions, dimensional consistency, conservation structure, known counterexamples. Verification must be reproducible outside the generating model (computer algebra, numerical spectral test, or a second model given the candidate but not the generation transcript). Survivors are promoted from \emph{candidate correspondence} to \emph{verified equivalence} and reported with their imported consequences; if nothing survives, the system \textbf{abstains}. Each verified correspondence enriches the representational repertoire available for subsequent retrieval, making the architecture recursively self-improving---in its search substrate, we note, not its reasoning engine: what grows with each cycle is the space of representations and known equivalences over which the next cycle operates.

\subsection{Design principles}

\emph{Two objectives, never blended.} Stage A is rewarded for relevant serendipity---operationally, candidate quality scored by structural-match depth net of surface similarity, weighted by domain distance and richness of the prospective import; Stage B is rewarded for rejection. Generate candidates aggressively; verify them mercilessly. The system-level currency is precision among \emph{assertions}; internal candidate volume is a free parameter.

\emph{Abstention as default.} Most candidates should die. Expected failure modes are stated in advance: decorative figures carry no structure-forced conventions; conventions occasionally collide (an arrow may denote a vector, a map, or a cause); distinct systems can share a visual gestalt; and where a twin's source supplies no computable invariant, verification is impossible and abstention mandatory. A well-functioning loop frequently reports ``no verified cross-domain identity''---abstention is not failure; it is the price of assertion meaning something.

\emph{Ranking.} Candidates are ordered by (i) structural-match depth net of surface similarity, (ii) verifiability---does a computable invariant exist?---and (iii) expected value of the import if true. Verification effort follows this order.

\emph{Scaffolded now, autonomous later.} In the motivating case the search strategy was human-supplied; the executed system is \emph{scaffolded abduction}. Whether the loop can originate its own search policy---when to invoke visual analogy, at what motif granularity, with what queries, and how to allocate verification---is an empirical hypothesis, not an engineering formality, and the evaluation program treats it as such.

\subsection{Interfaces}

Input: a source model (theory, dataset, formal system) plus its generated views. Output: either a scientific report (verified equivalence, dictionary, imported consequences, verification log) or a structured abstention (candidates considered, checks failed). Component responsibilities are separable, so the verifier can be formalized, strengthened, or replaced independently of the generator.

\section{What the Case Establishes---and Does Not}

The motivating case establishes: \emph{logical possibility} (a counterexample to ENT, provided one grants the episode instantiates identity abduction under scaffolding); \emph{mechanistic plausibility} (the operative cue is identifiable and is a drawing convention); \emph{architectural feasibility} (every step is a buildable component); and \emph{testable predictions} (Section~8). It does not establish: general intelligence; general scientific creativity; autonomous research capability; reliability at any frequency; or that the memory--lensing correspondence is novel to the joint literature---novelty is a claim about the discovery's value, is presently ``not located in our search'' rather than ``new,'' and is irrelevant to whether the recognition event occurred. Keeping these registers separate is the paper's discipline, and we ask reviewers to hold us to it.

\paragraph{What this architecture is not expected to solve.} The framework targets identity abduction over stable, convention-bearing representations, and its boundaries follow from that: it is not expected to generate \emph{mechanistic explanations} (proposing the process behind a phenomenon rather than an equivalence between formalisms); \emph{causal or interventional hypotheses}, which require reasoning about manipulation and counterfactual dependence that no static representation exposes; \emph{hypotheses in domains without stable representational conventions}, where convention space is empty by the theory's own account; or \emph{experimental designs}, which allocate physical action rather than representational search. These are not defects to be engineered away in later versions; they mark where the mechanism's preconditions fail, and where embodied or interventional accounts of discovery may retain their full force.

\section{Evaluation Program: The Diagram Abduction Benchmark (DAB-30)}

A mechanistic proposal requires a correspondingly mechanistic evaluation. Existing multimodal benchmarks assess diagram understanding, visual question answering, or domain-specific interpretation; none directly evaluates whether a system can \emph{generate, verify, and appropriately reject} cross-domain scientific identity hypotheses. We therefore propose the Diagram Abduction Benchmark (DAB-30) as the evaluation framework for representational grounding and the Abduction Loop. It is designed as living infrastructure: DAB-30 \emph{Version~1} comprises thirty benchmark instances, and the conceptual framework is independent of that count, admitting future expansion and domain-specific subsets without change.

\subsection{Objective and design}
DAB-30 answers a single question: can a system use scientific representations to generate and verify structurally meaningful cross-domain hypotheses, rather than merely interpret diagrams? The unit of evaluation is the complete abductive episode, not isolated recognition or retrieval. Instances are scientific figures spanning physics, mathematics, biology, chemistry, neuroscience, engineering, and Earth science, selected because they encode significant structural content through community-standard conventions---not because they are visually distinctive. Each instance is a single source figure presented without contextual explanation; systems receive no information about the originating discipline or any intended correspondence.

\subsection{Instance classes}
\begin{figure}[t]
\centering
\includegraphics[width=0.62\textwidth]{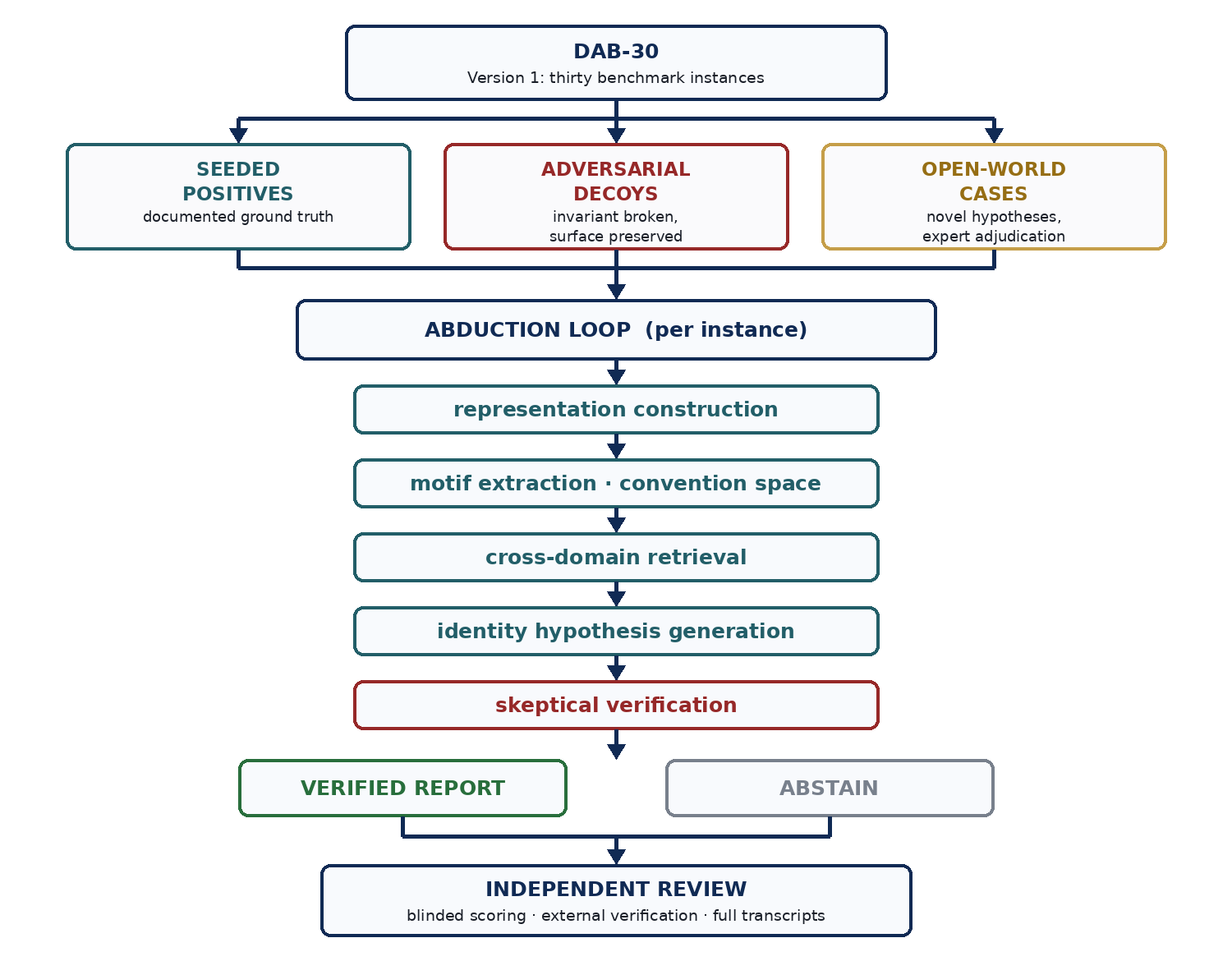}
\caption{DAB-30 structure. Each instance, from any of the three classes, is processed by the full Abduction Loop under blinded conditions; every outcome---verified report or abstention---passes to independent review.}
\label{fig:dab}
\end{figure}

Three complementary classes evaluate different facets of abduction (Figure~\ref{fig:dab}). \emph{Seeded positives}: figures with independently documented cross-domain correspondences, withheld from the system, providing objective ground truth for retrieval and verification. \emph{Adversarial decoys}: figures constructed or selected to preserve superficial visual similarity while disrupting a critical structural invariant; a sound system rejects these through explicit verification. \emph{Open-world cases}: previously unevaluated figures where plausible correspondences may exist but are not established, assessing novel-hypothesis generation under appropriate uncertainty.

\subsection{The abduction task}
For each figure the system must execute the full loop: construct alternative structural representations; decompose into motifs; retrieve candidate cross-domain analogues; formulate explicit identity hypotheses; attempt to falsify each against computable structural invariants; and either report a verified correspondence or explicitly abstain. Verification is mandatory---candidate generation alone does not constitute success.

\subsection{Evaluation dimensions and success criteria}
Performance is assessed on five independent dimensions: \emph{structural comprehension} (does the system identify the salient structural organization of the source?); \emph{representational abstraction} (the quality of the representation-independent structural description); \emph{cross-domain hypothesis generation} (relevance, diversity, and explanatory value of proposed correspondences); \emph{adversarial verification} (the rigor of deductive, mathematical, or computational checking before acceptance); and \emph{calibrated abstention} (rejection of unsupported correspondences when evidence is insufficient). The benchmark intentionally rewards conservative scientific behavior: false discoveries incur greater penalty than missed opportunities, so systems are judged on precision among accepted hypotheses, correctness of verification, confidence calibration, appropriate abstention, and reproducibility of supporting evidence. The intended operating point favors reliability over novelty---which is precisely the operating point of the architecture's Stage~B.

\subsection{Protocol, reproducibility, and intended use}
Runs occur under blinded conditions: fresh sessions with no conversational history, no benchmark-specific prompting beyond the published protocol, fixed and recorded model versions, complete transcript preservation, and independent verification outside the generating model whenever a hypothesis survives. Every run records model version, prompts, retrieved materials, intermediate hypotheses, verification artifacts, and final decisions, permitting full reconstruction of the reasoning episode by external reviewers. Because publication of any target contaminates all web-connected systems, runs on a target must precede circulation of that target; accordingly, the motivating case of Section~5---published here---is excluded from the seeded-positive arm, and all DAB-30 targets remain embargoed until their runs complete. DAB-30 does not measure general intelligence, expertise, or creativity; its purpose is narrower: to determine whether representational grounding and convention-space retrieval support reliable cross-domain scientific abduction under controlled conditions. Full operational detail (session hygiene, blinded dual scoring, preregistration, ablation conditions, complete metric suite) appears in Appendix~B; implementation artifacts (schemas, rubrics, prompts, execution scripts) belong to Paper~2 and the supplementary materials. We report no results here; Paper~2 is the benchmark's execution.

\section{Discussion}

\emph{For philosophy of science:} nothing here demotes embodiment. It remains a sufficient grounding route, the developmental origin of human representational competence, and the historical source of the conventions our mechanism exploits. What comes under pressure is only the necessity claim: grounding and embodiment are not synonyms, and the external-representation tradition has said so for forty years. \emph{For AI-for-science:} the proposal is architectural rather than scale-driven---the missing capability is not more parameters but a component structure that moves between representations and then submits to an adversarial verifier with the right to say no. And if representational grounding proves reliable, the implication is larger than one architecture: the dominant AI-for-science paradigm may shift from scaling ever-larger language models toward designing systems that deliberately generate, transform, compress, and compare representations \emph{before} reasoning over them---systems whose creativity is engineered at the level of representation, not emergent from parameter count. \emph{For retrieval:} convention-space indexing is a concrete, buildable answer to cross-vocabulary scientific search, and would be valuable even if every other claim in this paper failed. \emph{For scientific practice:} the account makes representation itself an active computational resource---communities that curate their conventions well are, on this view, building the retrieval infrastructure of future discovery without knowing it.

\paragraph{Scope.} The present benchmark evaluates abductive reasoning \emph{conditioned on the initiation of representational search}. Given a source representation, the agent is tasked with generating, evaluating, and appropriately calibrating candidate cross-domain correspondences; the benchmark therefore measures the quality of representational abstraction, hypothesis generation, adversarial evaluation, and calibrated abstention once candidate retrieval has begun. It does not estimate the base rate of spontaneous abductive insight in unconstrained environments, nor does it model the attentional or cognitive mechanisms by which a representation is initially selected for comparison.

\section{Limitations}

One motivating case, retrospective and imperfectly controlled (contamination channels enumerated in Appendix~A cannot be excluded post hoc); no prevalence, reliability, or autonomy claims; a human-supplied search strategy, making the executed system scaffolded; a diagram-first implementation of a mechanism claimed to be broader than diagrams; a benchmark designed but not yet executed; and normalization constants in the mathematical equivalence that are convention-dependent, stated explicitly here rather than deferred. Each limitation maps to a component of the evaluation program; none, we have argued, touches the logical point: a universal ``cannot'' now has a documented case that, if accepted as genuinely abductive, is its counterexample---and a benchmark on which the dispute over that acceptance can be settled.

\section{Future Work}

Execute DAB-30 with blinded cross-model runs and publish all transcripts (Paper~2). Complete the visual/textual ablation battery to isolate the operative channel. Formalize the verifier (computer-algebra integration; proof assistants where feasible). Extend representation construction beyond diagrams to other structure-exposing substrates---phase portraits, tensor networks, sonification is not excluded in principle. Score not only the discovery of correspondences but their promotion into shared formalisms, with measurable downstream yield: a correspondence that supports algorithm transfer and new predictions is worth more than one that does not. Test autonomous search-policy origination against the scaffolded baseline. Integrate literature-scale figure indexing so convention-space retrieval operates over the drawn corpus of science rather than ad hoc web search.

Test the \emph{projection-as-motif} conjecture. Where dimensional overflow fragments convention space (Section~4.2), the choice of projection may itself be structure-forced: one draws a Poincar\'e section when a return map exists, a bifurcation diagram when a parameter organizes the family, a recurrence plot when topological rather than metric structure is at issue. If so, convention space acquires a second layer indexed not by what was drawn but by what was discarded---two communities that shed the same degrees of freedom are asserting the same thing about which structure is essential. This is in principle a stronger twin signal than visual similarity, since it survives the mismatch that defeats the first layer, and it is continuous with the compression clause of Section~3. DAB-30 as specified does not test it; a dynamics-specific subset would.

\section{Conclusion}

We have not argued that embodiment is unnecessary for scientific reasoning in general, and we have not argued that current language models are scientists. We have argued something narrower and, we believe, more durable: representational transformations can expose structural invariants sufficient to ground at least some abductive scientific inferences; scientific diagrams, through their structure-forced conventions, provide one practical substrate; the Abduction Loop operationalizes the mechanism with divergent generation, adversarial verification, and principled abstention; and the DAB-30 program renders every load-bearing claim empirically decidable. Whether the mechanism is reliable is now an experimental question rather than a philosophical one---which is, we submit, exactly where a question about machine abduction belongs. Science has already drawn its own index; what remains is to build the systems that can read it.

\appendix
\section{The motivating case: protocol record and epistemic status}

\begin{figure}[h]
\centering
\includegraphics[width=0.92\textwidth]{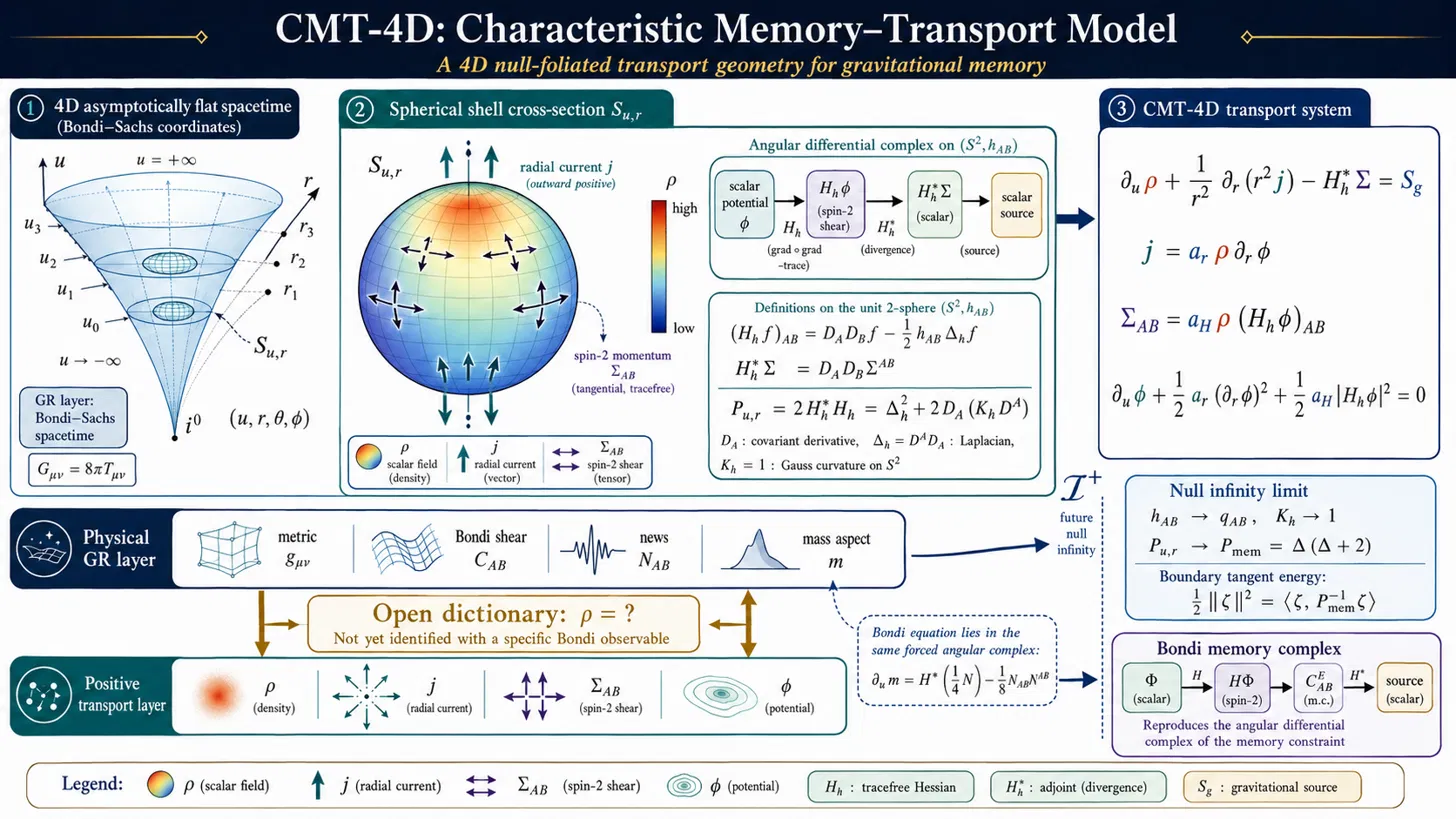}
\caption{The July 10, 2026 stimulus. The CMT-4D summary figure shown to the model, reproduced so that caveat~(b) can be assessed directly rather than taken on report: the labels and equations visible here are the embedded text whose contribution conditions~(ii)--(iv) of Appendix~B are designed to isolate. No target field was named, no cosmological context was supplied, and no correspondence was suggested; the accompanying instruction was to search the web for visually analogous images and examine their sources for possible model enhancements. The figure was itself generated from the mathematics, which is why representation construction appears as Stage~1 of the architecture rather than as a precondition of it.}
\label{fig:stimulus}
\end{figure}

The verbatim prompt, session conditions, and T1--T7 chronology follow the v0.x record of this program and are summarized here. Model: Claude Fable 5 (Anthropic), single session, July 10, 2026. Epistemic status: a discovery-generating observation, not a controlled experiment. Known caveats, stated in advance of any replication: (a) the session occurred within an ongoing research program with the same model; although no prior session mentions cosmology, lensing, or the CMB, retroactive exclusion of all contamination channels (account memory, retained context, indexed materials, file metadata) is impossible; (b) the input figure (Figure~\ref{fig:stimulus}) contains labels and equations, so the executed condition is figure-plus-embedded-text, and the operative channel is an open question for the ablation battery; (c) the search meta-strategy was human-supplied; and (d) the retrieved image layer of the session renders client-dependently and is therefore not byte-stable after the fact. The record of record for this paper is the mobile rendering retained by the author, which contains three result groups corresponding one-to-one with the three motif searches described in the transcript---spin-2 polarization on the sphere, weak-lensing shear and Kaiser--Squires mass mapping, and characteristic-extraction foliations---and is thus corroborated by the independent and stable textual layer rather than by repetition alone. A desktop rendering of the same session returned a truncated subset and is noted here only for completeness. We record this because it is a limitation of \emph{ad hoc} web retrieval as an evidentiary substrate, and because it is precisely the failure mode that DAB-30's frozen, hashed, source-traced corpus is designed to eliminate: in the benchmark, what was retrieved is fixed before the run and recoverable after it, which is not true of the episode reported here. The complete textual transcript and tool log will be released with the empirical companion paper, together with citations to the sources the model identified. The retrieved images themselves are third-party material and are described and cited rather than reproduced; per caveat~(d), the ranked image set is in any case not reconstructible with certainty after the session. The clean cross-model replication protocol (fresh incognito sessions, S1--S4 scoring, contamination probes, prompt-dependence control) is retained from the v0.x manuscript and is folded into DAB-30 as its seeded-positive arm; publication of this manuscript permanently contaminates the CMT-4D target; the motivating case is therefore \emph{excluded from the seeded-positive evaluation arm}, and DAB-30's seeded positives are distinct, embargoed targets whose runs precede any circulation of their descriptions. No inconsistency remains between the protocol and this paper's own publication.

\section{DAB-30: full protocol}
\emph{Session hygiene:} fresh sessions; memory, personalization, and custom instructions disabled; exact model identifiers and version strings logged; no hints beyond ``continue,'' five-turn cap.

\emph{Terminology.} Three roles are kept distinct throughout, because conflating them is the most likely way for a benchmark of this kind to overstate what it has shown. An \emph{internal skeptical reviewer} is a context-isolated review pass that may use the same model family; it detects overclaiming but does not constitute verification. An \emph{independent verifier} is a different model family, a computer-algebra or numerical check, a proof assistant, or a domain expert; only this satisfies the reproducibility requirement of Section~6.1. An \emph{independent human scorer} is a blinded evaluator of outputs, neither a generator nor a scientific verifier.

\emph{Multilevel motif extraction.} Motif extraction operates at four levels of increasing structural specificity, and the level attained bounds what may be claimed. \textbf{M1}, generic topology: tree, cycle, layered directed graph, hub-and-spoke, nesting. \textbf{M2}, typed relational grammar: causal direction, state transition, contraction, incidence, containment, composition, feedback, source--sink flow, quotient structure. \textbf{M3}, invariant-bearing convention: headlessness encoding $180^\circ$ symmetry, chirality, parity restriction, conserved node balance, kernel or null-space representation, operator order, commutative path equality, nesting as causal containment. \textbf{M4}, formal fingerprint: typed adjacency structure, symmetry group, spectrum, kernel dimension, operator composition, boundary-condition class, homology, conservation signature. M1 and M2 are shared with generic diagram understanding; only M3 and M4 test the convention-space mechanism of Section~4, since only they carry the invariants that make a convention structure-forced. This hierarchy is the operational form of the deep-versus-surface similarity gate named in Section~6.1.

\emph{Graded correspondence levels.} Outcomes are reported on a five-level scale rather than as a binary between analogy and identity, because the intermediate states are where most cross-domain correspondences legitimately sit and a binary decision rule forces them to one pole or the other. \textbf{L1}, structural analogy: a shared generic topology with no typed invariant preserved. \textbf{L2}, candidate homomorphism: at least one typed M2 relation preserved under an explicit partial mapping. \textbf{L3}, candidate isomorphism: at least one M3 invariant preserved, with a stated dictionary and no known violated invariant. \textbf{L4}, operational identity: an M4 fingerprint match together with an executed discriminating test. \textbf{L5}, replicated identity: L4 plus independent replication. Advancement is gated by motif level---an M1-only match is capped at L1 regardless of narrative plausibility, L2 requires an M2 match, L3 an M3 match, L4 an executed M4 test, L5 independent replication. Levels are floors on evidence, not confidence expressions.

\emph{Correspondence signature and Identity Readiness Score.} Each candidate carries a five-component signature: mapping completeness, structural-invariant preservation, operational correspondence, adversarial robustness, and predictive fidelity. In this paper the component vector is reported without aggregation; any aggregate Identity Readiness Score, together with its scale, weighting, and treatment of missing components, will be preregistered before benchmark execution rather than fixed here. The signature refines the five evaluation dimensions of Section~8.4 at the level of the individual candidate, where those dimensions apply to the episode; readiness is explicitly separated from empirical verification, so that a structurally well-prepared correspondence with no executed test is scored as such rather than promoted.

\emph{Evaluation-depth parity.} Verification depth must not vary with the provenance or the expected outcome of a candidate. Candidates attaining the same level receive the same verification treatment, and any case where equal depth cannot be completed is excluded from level-dependent comparisons and reported with the asymmetry stated. Without this rule, differences in measured quality can be produced entirely by differences in how hard each candidate was checked.

\emph{Scoring:} blinded dual scoring with vendor strings redacted; disagreements resolved by a third blind evaluator; inter-rater agreement reported; all runs reported, none excluded post hoc. Relevance judgments over retrieved material are made on pooled, deduplicated, shuffled candidate sets with channel, rank, and score metadata removed, so that rater inconsistency cannot manufacture differences between retrieval conditions.

\emph{Preregistration:} full protocol, figure set, decoy construction, motif and level rubrics, and analysis plan publicly timestamped (e.g., OSF) before the first run. Where instance counts within a class are small, results are reported as a case series with individual episodes narrated, and rates are secondary.

\emph{Conditions:} (i) full labeled figure; (ii) text-redacted figure; (iii) motif crops; (iv) neutral-token figure; (v) equations only; (vi) prose only; (vii) random-retrieval baseline; (viii) generic multimodal-embedding retrieval baseline (e.g., CLIP-class); (ix) symbolic-invariant fingerprint baseline (compute spectral or structural fingerprints of the source's operators and search the literature for matches, per Section~4.1); (x) shared-abstract-term retrieval baseline (query the highest-level term common to the candidate domains---e.g., ``spin-2 field''---and score precision of the returned set, testing directly whether a non-discriminating shared term suffices, per Section~4).

\emph{Metric suite:} precision among asserted equivalences; recall on seeded positives; distribution of attained levels L1--L5 and of motif levels M1--M4; Identity Readiness Score distributions; confidence calibration; calibrated abstention against decoys; retrieval precision at rank on raw pre-retention rankings; cost-normalized yield (per figure and per unit compute); time-to-first-verified-identification; expert-rated consequence; raw verified-over-generated reported, never optimized.

\emph{Independent verification:} every check supporting an L4 or L5 classification reproduced by computer algebra or numerical spectral test, and by a second model given the candidate but not the generation transcript.

\emph{Contamination:} post-run probe for prior knowledge of any target; neutral-prompt control to establish prompt-dependence; targets embargoed until their runs complete.

\section*{AI assistance statement}
This manuscript was prepared with substantial AI assistance, disclosed in full given the paper's subject. The motivating case of Section~5 is itself an episode with a multimodal model (Claude Fable~5, Anthropic), documented in Appendix~A. Beyond that case, large language models were used throughout drafting and revision: for adversarial review of the argument, identification of internal inconsistencies, and iterative refinement of the evaluation protocol of Appendix~B. Figure~1 was generated programmatically; the generating script is deposited at \url{https://doi.org/10.5281/zenodo.21776805}. Benchmark implementation and execution are handled through a different model family from the one in the motivating case---deliberately, to maintain separation between the system under discussion and the tooling used to evaluate it. All scientific claims, the mathematical verification of Section~5, and the final text are the author's responsibility.

\end{document}